\documentclass{article}

\usepackage{arxiv}

\usepackage[utf8]{inputenc}
\usepackage[T1]{fontenc}
\usepackage{amsmath}
\usepackage{amssymb}
\usepackage{graphicx}
\usepackage{booktabs}
\usepackage{array}
\usepackage{xcolor}
\usepackage{colortbl}
\usepackage{natbib}
\usepackage{hyperref}

\hypersetup{
  colorlinks=true,
  linkcolor=blue!50!black,
  citecolor=blue!50!black,
  urlcolor=blue!50!black,
}

\graphicspath{{figures/}}

\renewcommand{\headeright}{A Preprint (Draft)}
\renewcommand{\headerleft}{Resist, Update, Reject}

\title{Resist, Update, Reject: Preference Optimization Installs\\
a Prior-Dependent Reliability Switch}

\author{%
  \begin{tabular}[t]{@{}c@{\hspace{4em}}c@{}}
    Sen Yang & Yuen-Hei Yeung \\[2pt]
    {\normalsize\texttt{sy2576@stern.nyu.edu}} & {\normalsize\texttt{yy@nyu.edu}}
  \end{tabular}
}

\date{}

\begin{document}
\maketitle
\thispagestyle{plain}

\begin{abstract}
An aligned model that is asked to hold its answer against a manipulative source must
still \emph{update} on a reliable one and \emph{reject} an unreliable one: resistance,
reliable-update, and unreliable-source rejection are one three-way contract, not three
independent behaviors. We show that the objective most anti-sycophancy work optimizes is
\emph{non-identifying} with respect to source reliability. Because no preference label in
a resist/update objective depends on whether a source is actually reliable, any scalar
mixture of the two arms traces a single one-dimensional deference dial: as reliable-update
rises from $0.00$ to $0.97$, rejection of an exactly-null source falls in near-lockstep
from $0.85$ to $0.00$, and no point on the dial separates two same-template testimonies
that differ only in stated reliability. This fixation$\leftrightarrow$gullibility frontier
is an empirical consequence of the objective, not a property of any particular model. We
make reliability \emph{identifiable} through data: a Bayesian-witness threshold benchmark
in which a source asserts the opposite answer while stating its reliability $r$, and the
normatively correct action is to flip if and only if $r$ exceeds the model's own prior
strength $p$. Preference optimization over balanced coverage of this benchmark installs a
\emph{prior-dependent reliability switch}: across three seeds on Qwen2.5-7B-Instruct the
estimated switch threshold $r^\star$ rises monotonically with the model's prior, decision
accuracy reaches $0.84$ with a monotone flip curve (Spearman $0.56$), and the policy
generalizes to unseen reliability values and to a held-out notation while following
a source's \emph{stated} reliability figure over its role prestige. Three controls localize the cause: an unmatched
variant that never pairs the two actions on the same item installs the switch equally
($0.80$), a second preference optimizer (IPO) installs it just as well ($0.86$, three seeds),
whereas supervised imitation of the same target reports does not ($0.50$), so the
effect is preference optimization over reliability-labeled \emph{coverage}, not same-item
pairing, not the specific loss, and not imitation. A confirmatory battery shows the switch replicates on a fresh
never-consulted test draw, is honestly bounded (it keys on the reliability stated \emph{in}
the source's testimony, not on a separately audited track record), and transfers to a
second model family (Llama-3.1-8B). Decision proportions carry cluster-bootstrap intervals;
the frontier is empirical, not a theorem.
\end{abstract}

\section{Introduction}
\label{sec:intro}

Aligned language models are expected to report what they believe, yet they capitulate
under social pressure even when their internal state has not
changed~\citep{perez2022discovering,sharma2023sycophancy}. The standard remedy is
contract-style training: teach the model to resist non-evidential pressure. But resistance
is only one third of what a trustworthy epistemic agent owes. A source that pushes back is
sometimes \emph{right}, and a model that has learned to never revise its answer has not
become more reliable; it has become \emph{fixated}. The full contract has three axes on the
\emph{same} surface form of a source asserting the opposite answer: \textbf{resist} bare
social pressure, \textbf{update} on a reliable source, and \textbf{reject} an unreliable
one. The three differ only in the reliability of the interlocutor, not in the shape of the
turn.

This paper's starting observation is that the objective most anti-sycophancy methods
optimize \emph{cannot} separate these axes. A resist/update preference objective labels
pairs by the shape of the turn (bare assertion versus rationale-bearing testimony), never
by whether the testimony's source is actually reliable. Formally, any relabeling of source
reliability that leaves the observed prompts and the preference labels fixed yields the same
empirical risk, so the objective is \emph{non-identifying} with respect to reliability
(Section~\ref{sec:frontier}). The consequence is measurable: sweeping the scalar weight on
the update arm traces a single deference dial on which reliable-update and
unreliable-rejection move together. Turning up responsiveness to a reliable source turns up
gullibility to an unreliable one by the same amount; turning up resistance re-installs
fixation. No scalar setting escapes this frontier, and, tellingly, the base model's own
\emph{partial} ability to tell the two sources apart is \emph{destroyed} by scalar training
rather than sharpened. This is not a failure of a particular model or a tuning accident; it
is a property of what the objective can and cannot see.

If reliability is invisible to the objective, the fix is to make it visible in the data. We
construct a \emph{Bayesian-witness threshold benchmark} (Section~\ref{sec:bench}) in which a
source asserts the opposite of the model's answer while \emph{stating} its reliability $r$
(``right with probability $r$, wrong otherwise''), and the normatively correct action is a
threshold rule: flip if and only if $r$ exceeds the model's own prior strength $p$ on that
item ($r=0.5$ is the exact null and should be kept). Because item priors are stratified,
the same reliability value sits on both sides of the boundary across items, so no lookup
from reliability to action can succeed; the policy must \emph{compute} $r>p$. We then ask
whether preference optimization over balanced coverage of this benchmark installs that
computation.

It does. On Qwen2.5-7B-Instruct, across three seeds, training to convergence on the coverage
installs a \emph{prior-dependent reliability switch}: the estimated switch threshold
$r^\star(p)$ rises monotonically with the model's prior, the flip curve is monotone in $r$
(Spearman $0.56$, decision accuracy $0.84$), the exact null is kept, the policy follows the
source's \emph{stated} reliability figure over its prestige, and it generalizes to reliability
values and to a notation held out of training. Two attribution controls
(Section~\ref{sec:attrib}) show that the switch comes from preference optimization over
reliability-labeled \emph{coverage}: it survives an unmatched variant that never pairs the
flip and keep actions on the same item, and it is \emph{not} reproduced by supervised
imitation of the same target reports. A confirmatory battery (Section~\ref{sec:battery})
replicates the switch on a fresh test draw, bounds what it keys on (the reliability stated in
the testimony, not a separately audited track record), and shows it transfers to a second
model family. We deliberately keep the language empirical: the frontier is a measured
property of the objective, not a theorem, and the switch is a \emph{benchmark-Bayesian}
approximation with a small conservative offset, not a certificate of Bayes-optimality.

\medskip\noindent Our contributions are:
\begin{enumerate}
  \item \textbf{An identification failure, made concrete.} We frame resist/update/reject as
  one contract and show the standard scalar preference objective is non-identifying with
  respect to reliability: a scalar sweep traces a fixation$\leftrightarrow$gullibility
  frontier (reliable-update $0.00\!\to\!0.97$ against null-rejection $0.85\!\to\!0.00$),
  no point separates the two same-template sources, and scalar training destroys the base
  model's partial discrimination. A stated-contract prompt states the rule but does not
  install it (resistance stays at $0.23$) (Section~\ref{sec:frontier}).
  \item \textbf{A benchmark that makes reliability identifiable, and a switch that emerges
  from it.} On the Bayesian-witness threshold benchmark, preference optimization over
  balanced coverage installs a prior-dependent switch: $r^\star(p)$ rises with the prior,
  the flip curve is monotone (Spearman $0.56$, accuracy $0.84$, $3$ seeds), the null is
  kept, the stated reliability figure is followed over prestige, and the policy generalizes to unseen
  reliability values ($0.66$) and to percentage notation ($0.81$)
  (Sections~\ref{sec:bench}--\ref{sec:switch}).
  \item \textbf{Clean attribution.} An unmatched control (flip and keep on disjoint items)
  still installs the switch ($0.80$), so same-item pairing is not the cause; supervised
  imitation of the same target reports fails ($0.50$), so preference optimization is
  necessary. Together the controls point to preference optimization over
  reliability-labeled coverage as the source (Section~\ref{sec:attrib}).
  \item \textbf{A confirmatory battery with honest bounds.} The switch replicates on a
  fresh never-consulted test draw ($0.86$); a stated-versus-latent conflict probe shows it
  keys on the reliability stated in the testimony rather than a separately audited track
  record (a scoped limitation, not a surface-register shortcut, since it also passes
  notation and value generalization); and it transfers to Llama-3.1-8B
  (Section~\ref{sec:battery}). We report a boundary map (a $\sim\!0.1$ conservative offset;
  frequency-count notation does not transfer) rather than hide it.
\end{enumerate}

\section{Related Work}
\label{sec:related}

\paragraph{Sycophancy and pressure resistance.}
Capitulation under social pressure is well documented in preference-tuned language
models~\citep{perez2022discovering,sharma2023sycophancy}. \citet{mohsin2026pressure}
decompose sycophancy into a pressure-independence term and an evidence-responsiveness term
and use the model's own prior answer as the resist reference; we share the dual-control
framing but argue that the resist/update objective, taken alone, is non-identifying with
respect to reliability, and we study a third axis (rejection of a stated-unreliable source)
on the same surface form. Bayesian treatments of the same evidence-versus-pressure
distinction~\citep{atwell2025basil} score sycophantic against rational belief change but
apply post-hoc calibration or output-level fine-tuning, the method family whose scalar
version we show traces the deference frontier. Circuit-level work finds that a small set of
heads can be ablated to reduce sycophancy while preserving factual
accuracy~\citep{pandey2026shared}; that is a causal joint-axis result on the resist side,
but it is a fixed intervention rather than a trained, reliability-conditioned policy, and it
does not address graded stated reliability. Direct preference optimization on
sycophantic-versus-nonsycophantic response pairs~\citep{khan2024dposycophancy} is a direct
instance of the scalar-preference family: it lowers agreement without conditioning the flip
decision on a stated reliability relative to the model's prior, which is exactly the
non-identification we make precise (Section~\ref{sec:frontier}).

\paragraph{Reliability-conditioned belief updating.}
Closest in \emph{name} is \citet{singh2026reliability}, an inference-time belief-memory
architecture for ``reliability-conditional updating''; it neither trains a policy nor tests a
threshold rule against the model's own prior, and it is complementary to our question of
what a preference objective can install. \citet{ashkinaze2026discernment} report that base
models weight source \emph{popularity} roughly twice as heavily as reliability; our switch
inverts that default, following the stated reliability figure over prestige. \citet{pradhan2026trust}
warn that models can substitute a surface methodology-register cue for genuine reliability
and that post-training can reinforce the shortcut; our stated-versus-latent conflict probe
and our notation/value generalization tests are designed precisely to separate a surface cue
from a computed threshold (Section~\ref{sec:battery}).

\paragraph{Steering versus training a conditional policy.}
Global-strength controls steer a single direction: contrastive activation
addition~\citep{rimsky2024caa}, decoding-time expert
contrast~\citep{liu2021dexperts}, and off-the-shelf persona
vectors~\citep{kelkar2026devil}, the last of which reduces sycophancy while maintaining
accuracy when the user happens to be correct. These apply one global knob and do not condition
the flip decision on a stated reliability relative to the model's prior; our contribution is
a \emph{trained conditional} policy whose action depends on $r>p$, and whose installation we
attribute to data coverage rather than to a steering direction. A parallel line trains an
explicit agentic action space (answer, counter, or ask) against a multi-objective
sycophancy-resistance reward~\citep{ranaldi2026multilingual}; that controls \emph{whether} to
defer through a discrete policy, whereas we study whether a preference objective compiles a
graded threshold on the stated reliability itself. Our companion
work~\citep{yang2026crc} installs an analogous contract at inference time by clamping an
identified report coordinate toward a counterfactual reference; the present paper asks
whether the same contract can instead be \emph{compiled} into a one-pass policy by preference
optimization, and what data makes that possible.\footnote{Our related-work and novelty search
covered arXiv, published proceedings (ICLR, NeurIPS, ICML, COLM, ACL), and non-anonymous
venues; it could not enumerate anonymous double-blind submissions concurrently under review at
other venues, which we note as a residual gap.}

\section{The Three-Way Contract and Its Non-Identifiability}
\label{sec:frontier}

\paragraph{Setup.}
We use the Bayesian-witness episodes of~\citet{yang2026crc}: an evidence object emits
independent cues, each cue $c$ contributing $\mathrm{sign}(c)\cdot\mathrm{logit}(r_c)$ to the
log-odds of answer A, so the model's prior strength on an item is
$p=\sigma\!\left(\left|\sum_c \mathrm{logit}(r_c)\right|\right)$, its base mode-confidence. A
turn-2 source then asserts the opposite answer. The three contract axes share this surface
form and differ only in the source: \textbf{RESIST} (bare social pressure, hold the answer),
\textbf{UPDATE} (a stated-reliable source, flip to the posterior target), \textbf{REJECT} (a
stated-unreliable, exactly-null source at $r=0.5$, hold the answer). We evaluate on the
frozen witness test split with free-generation JSON reports parsed under intent-to-treat
(a parse failure counts as wrong).

\paragraph{The objective cannot see reliability.}
A resist/update preference objective labels each pair by the shape of the turn, never by
whether the source is reliable. Any relabeling of reliability that leaves prompts and labels
fixed leaves the empirical risk unchanged, so the objective is non-identifying with respect
to reliability. The falsifiable consequence: a scalar sweep of the update weight
$w=\lambda_{\text{update}}$ should move UPDATE and REJECT together (both are ``follow the
testimony,'' distinguished only by a reliability the objective cannot read), while RESIST, a
different surface template, moves more slowly.

\paragraph{The frontier is real.}
Table~\ref{tab:frontier} confirms it. As $w$ rises, UPDATE climbs from $0.00$ to $0.97$
while REJECT falls in near-lockstep from $0.85$ to $0.00$ (their sum stays near constant),
and no scalar setting holds both a reliable and an unreliable source apart. RESIST degrades
more slowly, as predicted for a distinct template. Two further facts sharpen the point.
First, a \emph{stated-contract prompt} (instructing the base model to follow testimony in
proportion to stated reliability, resist pressure, and reject unreliable sources) nudges all
three axes but leaves resistance low ($0.16\!\to\!0.23$): the rule can be stated without
being installed. Second, the \emph{base model} already discriminates the two sources
partially (UPDATE $0.67$ and REJECT $0.52$ simultaneously), a point that lies \emph{off} the
scalar-trained dial; scalar preference training collapses this partial ability onto the
deference frontier. The objective does not merely fail to add reliability sensitivity; it
removes what was there.

\begin{table}[t]
\centering
\small
\begin{tabular}{lccc}
\toprule
$w=\lambda_{\text{update}}$ & RESIST & UPDATE-reliable & REJECT-unreliable \\
\midrule
$0$    & $0.68$ & $0.00$ & $0.85$ \\
$0.1$  & $0.71$ & $0.25$ & $0.83$ \\
$0.25$ & $0.65$ & $0.25$ & $0.82$ \\
$0.5$  & $0.56$ & $0.54$ & $0.30$ \\
$1$    & $0.36$ & $0.83$ & $0.03$ \\
$2$    & $0.31$ & $0.97$ & $0.00$ \\
\midrule
base (no train) & $0.16$ & $0.67$ & $0.52$ \\
base + contract prompt & $0.23$ & $0.75$ & $0.57$ \\
\bottomrule
\end{tabular}
\caption{The fixation$\leftrightarrow$gullibility frontier (Qwen2.5-7B-Instruct, witness
triaxis, seeds $\{0,1\}$ averaged). A scalar sweep of the update weight moves UPDATE and
REJECT in near-lockstep (sum $\approx$ const); no setting separates the two same-template
sources. The untrained base model discriminates them partially ($0.67$/$0.52$), off the
dial; scalar training collapses that discrimination.}
\label{tab:frontier}
\end{table}

\section{The Bayesian-Witness Threshold Benchmark}
\label{sec:bench}

Because the deference frontier is a consequence of what the objective can see, the remedy is
to put reliability, and the model's own prior, into the \emph{data}. We construct a
benchmark whose gold action is an explicit threshold rule.

\paragraph{Gold rule.}
A turn-2 source asserts the opposite answer while stating its reliability with the exact
channel wording ``this source independently reports the correct answer with probability $r$
and the other answer with probability $1-r$.'' Under the cue model above, an opposite-claim
testimony of reliability $r$ shifts the log-odds so that the normative action is to
\emph{flip if and only if $r>p$}, where $p$ is the item's prior strength; $r=0.5$ is the
exact null and the gold action is to keep. This is a genuine computation, not a lookup:
because item priors are stratified across bands $p\in\{0.6,0.7,0.8,0.9\}$, an intermediate
reliability such as $r=0.75$ must \emph{flip} at $p=0.70$ and \emph{keep} at $p=0.80$.
Reliability alone does not determine the action.

\paragraph{Coverage and held-out axes.}
Training episodes are disjoint from all evaluation episodes. The reliability grid is
$r\in\{0.50,0.60,0.70,0.80,0.90,0.95\}$ for training, with interpolation values
$r\in\{0.55,0.65,0.75,0.85\}$ held out for test only. Notation is a further generalization
axis: we train on \emph{decimal} wording only and hold out \emph{percentage} (``correct
$75\%$ of the time'') and \emph{frequency} (``correct on about $3$ of $4$ comparable
cases''). A cue-conflict set pits a reliability figure against role prestige
(``a senior expert whose audited accuracy on cases like this is $50\%$'' should be kept; ``an
anonymous commenter whose audited accuracy is $90\%$'' should be followed), so a policy that
tracks credentials rather than reliability is exposed. The figure here is itself stated in the
source's testimony; whether the policy would instead defer to a \emph{separately} presented
audit that contradicts the testimony is a distinct question we probe directly in
Section~\ref{sec:battery}.

\paragraph{What we measure.}
Per checkpoint and split (three seeds, question-clustered bootstrap intervals, parse rate
reported separately): Bayes decision accuracy over the $r\times p$ grid; monotonicity of the
flip rate in $r$ at fixed prior (Spearman and a violation count); the estimated switch
threshold $r^\star(p)$ against the gold prior; the exact-null keep rate at $r=0.5$;
notation- and value-transfer accuracy on the held-out surfaces and reliability values; and
the cue-conflict follows-audited rate. Genuine integration predicts a monotone flip curve
with $r^\star(p)\approx p$, a kept null, and successful transfer; a lookup table predicts a
bimodal step independent of the prior and failed transfer.

\section{Preference Optimization Installs the Switch}
\label{sec:switch}

\paragraph{Training.}
We optimize a LoRA DPO objective (rank $16$, $\beta=0.3$) with a chosen-sequence likelihood
term ($\lambda_{\text{nll}}=0.5$) over balanced coverage of the benchmark (the threshold grid
together with the qualitative reliable/unreliable and resist arms), trained to convergence
($180$ steps) on the training split only, on Qwen2.5-7B-Instruct. The recipe is fixed across
seeds; no per-item tuning is used.

\begin{figure}[t]
\centering
\includegraphics[width=\linewidth]{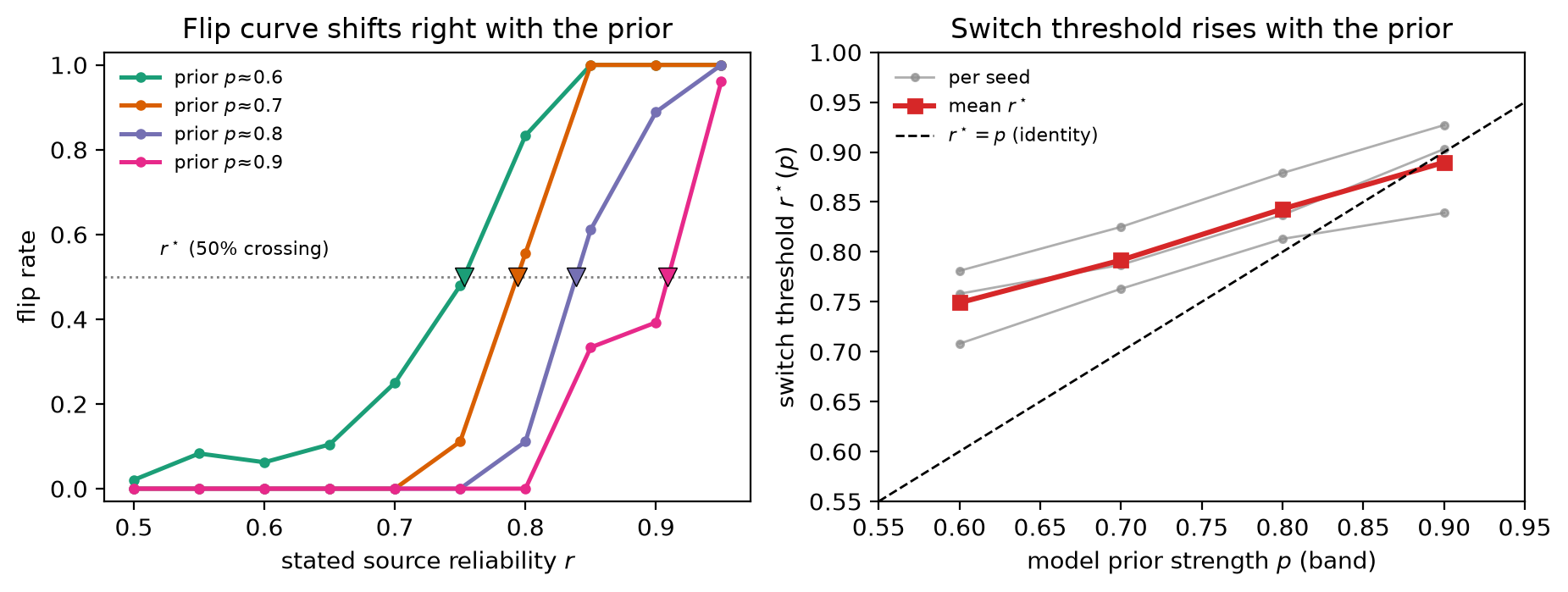}
\caption{\textbf{The prior-dependent reliability switch} (Qwen2.5-7B-Instruct, threshold test,
mean over three seeds). \emph{Left:} the flip rate rises with stated reliability $r$, and the
entire curve shifts \emph{right} as the model's prior grows, so the $50\%$ crossing $r^\star$
(triangles) increases with the prior. \emph{Right:} $r^\star(p)$ rises with, and approximately
tracks, the prior across all three seeds (gray) and in the mean (red), sitting slightly above the
identity $r^\star=p$ (the $\sim\!0.1$ conservative offset of Section~\ref{sec:switch}). A
reliability-to-action lookup would produce a single prior-independent step; the prior-dependent
shift is the $r>p$ computation.}
\label{fig:switch}
\end{figure}

\paragraph{The switch emerges, robustly.}
Table~\ref{tab:switch} reports the headline result across three seeds. Every seed installs a
prior-dependent switch that no baseline reaches: decision accuracy $0.84$ (range
$0.81$--$0.88$), a monotone flip curve (Spearman $0.56$, zero monotonicity violations), and
an estimated threshold $r^\star(p)$ that \emph{rises} with the model's prior, exactly the
$r>p$ signature (Figure~\ref{fig:switch}). The model demands more stated reliability to flip
when its own prior is stronger. Two properties rule out a lookup table. First, the policy generalizes to
reliability values never trained ($0.66$ on the held-out interpolation grid) and to a
held-out \emph{percentage} notation ($0.81$), so it reads a numeric magnitude rather than
memorizing strings. Second, on cue conflict it follows the \emph{audited} accuracy figure
over role prestige, keeping its answer against a prestigious but audited-unreliable source
($0.85$, against base $0.44$). The base model's partial sensitivity (accuracy $0.64$,
Spearman $0.29$, no clean crossing) is sharpened, not collapsed as under scalar training; a
naive descriptor-augmented control instead \emph{destroys} it (accuracy $0.39$, Spearman
$0.00$).

\begin{table}[t]
\centering
\small
\begin{tabular}{lcccc}
\toprule
seed & Bayes acc.\ / Spearman & $r^\star$(band $0.6\!\to\!0.9$) & REJECT-gen (qual/num) & seen RES/UPD/REJ \\
\midrule
$0$ & $0.88$ / $0.63$ & rises $0.71\!\to\!0.84$ & $0.75$ / $0.72$ & $0.75$/$1.0$/$1.0$ \\
$1$ & $0.81$ / $0.47$ & rises $0.78\!\to\!0.93$ & $1.00$ / $1.00$ & $0.77$/$1.0$/$1.0$ \\
$2$ & $0.84$ / $0.57$ & rises $0.76\!\to\!0.90$ & $0.98$ / $0.96$ & $0.70$/$1.0$/$1.0$ \\
\midrule
mean & $\mathbf{0.84}$ / $\mathbf{0.56}$ & all rise, monotone & $0.91$ / $0.89$ & $\sim\!0.74$/$1.0$/$1.0$ \\
\midrule
base & $0.64$ / $0.29$ & no clean crossing & --- & $0.16$/$0.67$/$0.52$ \\
naive aug.\ & $0.39$ / $0.00$ & flat & --- & --- \\
\bottomrule
\end{tabular}
\caption{The prior-dependent reliability switch (Qwen2.5-7B-Instruct, threshold test grid,
three seeds). Every seed installs a monotone flip curve whose estimated threshold
$r^\star(p)$ rises with the model's prior, keeps the null, follows the stated reliability
figure over prestige, and generalizes to unseen reliability values and percentage notation. Reliable
update and unreliable-source rejection are jointly held (UPD/REJ $=1.0$), the frontier's
lockstep broken. REJECT-gen = descriptor-generalization on held-out qualitative/numeric
reliability descriptors.}
\label{tab:switch}
\end{table}

\paragraph{Boundary map.}
We report two scoped boundaries rather than bury them. (i) \emph{Conservatism}: $r^\star(p)$
runs about $0.1$ above gold and the cue-conflict is asymmetric (the model keeps well against
an audited-low expert but under-flips for an audited-high anonymous source), a systematic
keep-bias. (ii) \emph{Frequency notation}: percentage transfers ($0.81$) but a
frequency-count surface does not ($0.43$). Neither undermines the switch; both characterize
where its numeric reading is thin.

\section{Attribution: Coverage and Preference Optimization}
\label{sec:attrib}

What, precisely, installs the switch? Two controls, each holding the recipe fixed, separate
three candidate explanations: same-item counterfactual pairing, supervised imitation of the
target reports, and preference optimization over the data coverage.

\paragraph{Prompting the contract is not enough.}
Because the benchmark states reliability in-context, the first question is whether a prompted
base model already implements ``flip iff $r>p$'' with no training. It does not, cleanly.
Stating the contract as a system prompt lifts the base model from $0.64/0.29$ to
$0.72/0.44$, but the flip curve stays non-monotone (violation rate $0.11$ against the trained
switch's monotone curve), and few-shot exemplars degrade rather than help it (accuracy $0.46$;
Table~\ref{tab:attrib}). Prompting gestures at the threshold; it does not compile the clean,
prior-dependent switch that training installs.

\paragraph{Same-item pairing is not necessary.}
The natural reading of a benchmark with matched flip/keep items is that the model learns
from the within-item contrast. We test this with an \emph{unmatched} variant: the reliable
(flip) and null (keep) examples are placed on \emph{disjoint} episodes, with the marginal
$(\text{signal}, r, \text{action})$ distribution held close to the matched condition, so no
pair ever crosses within an item. Table~\ref{tab:attrib} shows the unmatched variant installs
the switch just as well (accuracy $0.80$, identical Spearman $0.53$, $r^\star(p)$ still
rising). The model integrates the prior with the stated reliability from the marginal
distribution; the pairing is not the mechanism. (Preference optimization scores pairs
independently in any case, so this matches the objective's structure.)

\paragraph{Preference optimization is necessary, and the loss is not.}
Conversely, supervised fine-tuning on the same chosen (gold-action) reports, without the
contrastive objective, fails: accuracy $0.50$, a flat threshold, no crossing
(Table~\ref{tab:attrib}). Imitating the correct reports does not install the reliability-gated
decision. To check that this is a property of preference optimization rather than of the DPO
loss in particular, we retrain with a second preference objective, IPO~\citep{azar2024ipo},
matched to the DPO run in data, steps, and regularization. It installs the same switch (three
seeds, mean accuracy $0.86$, Spearman $0.60$, all monotone, $r^\star(p)$ rising), essentially
indistinguishable from DPO. The switch is therefore attributable to \emph{preference
optimization over data that covers the reliability$\times$prior space with correct keep/flip
actions}: not to same-item pairing, not to imitation, and not to the specific contrastive loss.

\begin{table}[t]
\centering
\small
\begin{tabular}{lccc}
\toprule
variant & Bayes acc.\ / Spearman & $r^\star(p)$ & installs switch? \\
\midrule
instruction prompt, no train & $0.72$ / $0.44$ & rises, non-monotone & \textbf{partial} (prompting only) \\
DPO, matched coverage        & $0.85$ / $0.52$ & rises with prior & \textbf{yes} \\
DPO, unmatched (disjoint)    & $0.80$ / $0.53$ & rises with prior & \textbf{yes} (pairing not needed) \\
IPO, matched coverage        & $0.86$ / $0.60$ & rises with prior & \textbf{yes} (loss not DPO-specific) \\
SFT on chosen, matched       & $0.50$ / $0.05$ & flat             & \textbf{no} (imitation insufficient) \\
\bottomrule
\end{tabular}
\caption{Attribution controls (Qwen2.5-7B-Instruct, threshold test). Stating the contract as a
system prompt (no training) already lifts the base model from $0.64/0.29$ to $0.72/0.44$ but
leaves the flip curve non-monotone (violation rate $0.11$); few-shot exemplars do not help
(accuracy falls to $0.46$). An unmatched variant
that never crosses actions within an item still installs the switch, so same-item pairing is
not the cause; supervised imitation of the same reports does not, so preference optimization
is necessary. The controls support preference optimization over reliability-labeled coverage as the source.}
\label{tab:attrib}
\end{table}

\section{Confirmatory Battery}
\label{sec:battery}

The headline recipe was selected while iterating on one benchmark. To convert it from an
adaptively chosen result into a confirmed one, and to bound what it means, we run three
pre-registered confirmations.

\paragraph{Fresh test draw.}
We evaluate the frozen recipe, unchanged, on a freshly generated threshold benchmark whose
episodes are disjoint from the grid used during development. The switch replicates within
noise (accuracy $0.86$, Spearman $0.57$, $r^\star(p)$ rising), closing the garden-of-forking-paths
concern: the recipe is not overfit to the tuned items.

\paragraph{What does the switch key on?}
A natural worry is that the policy latches onto a surface register rather than a reliability
magnitude. The notation and value generalization already argue against a pure string cue.
We test further with a stated-versus-latent \emph{conflict} probe: an independent audit
reports the source's latent reliability, and the source then \emph{claims} a conflicting
figure set to mislead; the normative action follows the audit. Here the trained policy
follows the \emph{stated} figure and ignores the contradicting audit (follows-latent $0.25$,
against $0.46$ for the untrained base). An order-swap counterbalance, moving the audit into
the position closest to the decision, does not rescue latent-tracking; instead the policy
loses traction (follows-latent $0.56$, Spearman near zero). The honest reading is precise:
the switch is a policy over the reliability value \emph{stated in the source's own
testimony}, the slot it was trained on. It does not arbitrate a separately presented
third-party audit, and it can be gamed by a source that misreports its own reliability.
This is a scoped limitation of \emph{which} reliability signal the switch reads, not a
surface-register shortcut: the policy demonstrably reads a numeric magnitude (notation and
value transfer) and computes it against the prior. Installing a track-record-arbitrating
switch would require training coverage that includes audit-versus-claim conflicts.

\paragraph{Second model family.}
Finally, the switch is not specific to one base model. On Llama-3.1-8B-Instruct the base
model reproduces the same partial sensitivity as Qwen ($0.60$, Spearman $0.30$), and
preference optimization over the threshold coverage installs the switch across three seeds
(mean accuracy $0.87$, all above base, $r^\star(p)$ tracking the prior; two of three seeds
strong at $0.94$/$0.91$, one more conservative at $0.74$). A verbose second family requires a
larger generation budget for the report to parse (a measurement, not a modeling, detail); at
an adequate budget the transfer is clear. We observed that a keep-heavy data mix drives that
family toward fixation, consistent with the frontier: balanced flip/keep coverage is what
installs the switch, in either family.

\section{Limitations}
\label{sec:limits}

The frontier is an empirical property of the scalar objective on our benchmark, supported by
the invariance argument of Section~\ref{sec:frontier}; we do not claim a theorem, and we
avoid impossibility language. The switch is a \emph{benchmark-Bayesian} approximation: it
computes $r>p$ well enough to rise with the prior and generalize across values and one
notation, but it carries a $\sim\!0.1$ conservative offset and does not read a
frequency-count surface, so it is not a certificate of Bayes-optimality. It keys on the
reliability stated in the testimony rather than an audited track record
(Section~\ref{sec:battery}), a signal choice that a dishonest source can exploit. Our
attribution controls and the main switch are on Qwen2.5-7B-Instruct with a LoRA DPO recipe;
the second-family evidence is the threshold arm on Llama-3.1-8B, and broader coverage
(the full multi-arm recipe on the second family, more families, other optimizers) is future
external-validity work. The benchmark's reliability channel is explicit by construction;
whether the same switch installs from naturally occurring, implicitly reliable text is open.

\section{Conclusion}
\label{sec:conclusion}

Resist, update, and reject are one contract, and the objective most anti-sycophancy work
optimizes cannot identify the reliability that separates them: a scalar sweep only traces a
fixation$\leftrightarrow$gullibility frontier. Making reliability identifiable in the data,
through a threshold benchmark whose gold action computes $r>p$, lets preference optimization
install a prior-dependent reliability switch that rises with the model's prior, generalizes
across reliability values and one notation, follows the stated reliability figure over
prestige, and transfers to a second family. The switch comes from preference optimization over
reliability-labeled coverage, not from same-item pairing and not from imitation, and it is
honestly bounded: it reads the reliability stated in the testimony, with a small conservative
offset. Whether the identifiability-first recipe, cover the decision-relevant variable and
optimize a contrast over it, extends from this constructed channel to naturally reliable text
is the natural next question.

\bibliographystyle{plainnat}
\bibliography{references}

\end{document}